\PassOptionsToPackage{table,xcdraw}{xcolor}
\documentclass{article} 
\usepackage{iclr2024_arxiv,times}

\usepackage{amsmath,amsfonts,bm}

\def\eqref#1{equation~\ref{#1}}

\def\1{\bm{1}}

\DeclareMathAlphabet{\mathsfit}{\encodingdefault}{\sfdefault}{m}{sl}
\SetMathAlphabet{\mathsfit}{bold}{\encodingdefault}{\sfdefault}{bx}{n}

\iclrfinalcopy

\usepackage[pagebackref,breaklinks,colorlinks]{hyperref}
\usepackage{url}
\usepackage{amsmath}
\usepackage{amssymb}
\usepackage{booktabs}
\usepackage{tabularx}
\usepackage{url}       
\usepackage{amsfonts}       
\usepackage{nicefrac}       
\usepackage{pifont} 
\usepackage[table,xcdraw]{xcolor}
\usepackage{enumitem} 
\usepackage{standalone}
\usepackage{tikz}
\usepackage{wrapfig}

\newcommand{\xmark}{\ding{55}}%
\newcommand{\cmark}{\ding{51}}%

\definecolor{TableGray1}{HTML}{9B9B9B}
\definecolor{TableGray2}{HTML}{C0C0C0}
\definecolor{TableGray3}{HTML}{EFEFEF}
\definecolor{TableGreen}{HTML}{BDDCAC}

\definecolor{rpevtarget}{HTML}{FF06FC}
\definecolor{rpevtp}{HTML}{FEFC30}
\definecolor{rpevtn}{HTML}{00F2E4}
\definecolor{rpevfp}{HTML}{4135A2}

\definecolor{ColCorrespondence}{HTML}{D883FF}
\definecolor{ColNavigSpace}{HTML}{F4B183}
\definecolor{ColExits}{HTML}{2F5597}

\newcolumntype{C}{>{\columncolor{TableGray2}}c}
\newcolumntype{L}{>{\columncolor{TableGray2}}l}

\newcommand\mydots{\makebox[0.7em][c]{.\hfil.\hfil.}}

\newcommand\myparagraph[1]{\textbf{#1} ---}

\title{End-to-End (Instance)-Image Goal Navigation through Correspondence as an Emergent Phenomenon}

\author{Guillaume Bono, Leonid Antsfeld, Boris Chidlovskii, Philippe Weinzaepfel, Christian Wolf \\
Naver Labs Europe\\
\texttt{\{firstname.lastname\}@naverlabs.com}
}

\begin{document}

\maketitle

\vspace{-0.2cm}

\begin{abstract}

\vspace{-0.1cm}

Most recent work in goal oriented visual navigation resorts to large-scale machine learning in simulated environments. The main challenge lies in learning compact representations generalizable to unseen environments and in learning high-capacity perception modules capable of reasoning on high-dimensional input. 
The latter is particularly difficult when the goal is not given as a category (``\textit{ObjectNav}'') but as an exemplar image (``\textit{ImageNav}''), as the perception module needs to learn a comparison strategy requiring to solve an underlying visual correspondence problem. This has been shown to be difficult from reward alone or with standard auxiliary tasks.
We address this problem through a sequence of two pretext tasks, which serve as a prior for what we argue is one of the main bottleneck in perception, extremely wide-baseline relative pose estimation and visibility prediction in complex scenes. The first pretext task, cross-view completion is a proxy for the underlying visual correspondence problem, while the second task addresses goal detection and finding directly. We propose a new dual encoder with a large-capacity binocular ViT  model and show that correspondence solutions naturally emerge from the training signals. Experiments show significant improvements and SOTA performance on the two benchmarks, \textit{ImageNav} and the \textit{Instance-ImageNav} variant, where camera intrinsics and height differ between observation and goal. 
\end{abstract}

\section{Introduction}
\label{sec:intro}

\noindent
\begin{wrapfigure}{r}{6cm} \centering
\vspace*{-4mm}
\includegraphics[width=\linewidth]{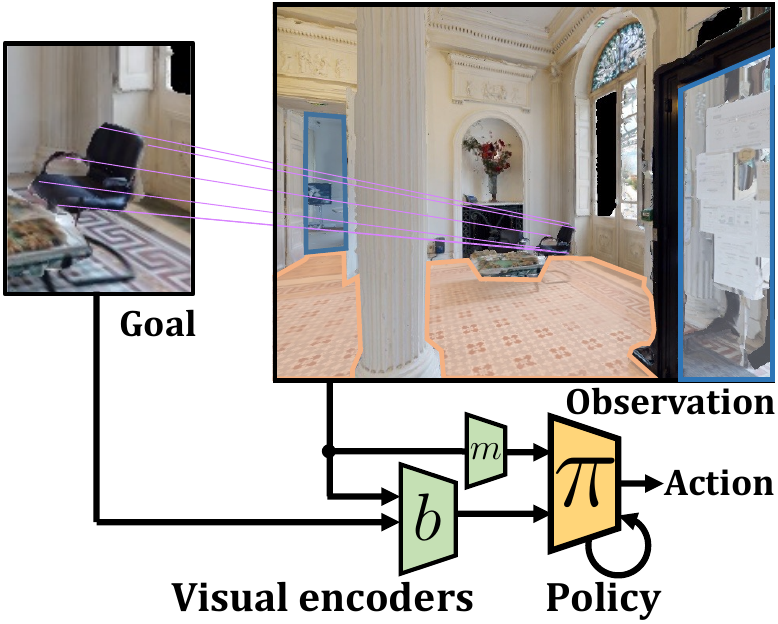}
\caption{\label{fig:teaser}Navigation skills include detecting \textcolor{ColNavigSpace}{\textbf{navigable space}},  \textcolor{ColExits}{\textbf{exits}}, and the agent's~\textcolor{ColCorrespondence} {\textbf{relative pose wrt. the goal}}. The correspondence solutions required by pose emerge from training with pretext tasks.}
\vspace*{-4mm}
\end{wrapfigure}
Goal oriented visual navigation is usually addressed through large-scale training in simulation, followed by sim2real transfer. While decision taking has not yet been solved either, recent research provides evidence that perception is a major bottleneck with several challenges:
learning representations required for planning;
extracting 3D information, difficult when depth is not available or not reliable; and, generalizing to unseen environments, which is challenging given the limited number of existing training environments.

The perception module of an agent needs to address several skills, which include detecting navigable space and obstacles, detecting exits necessary for long horizon planning, detecting goals and estimating the agent's relative pose with respect to them, see Figure~\ref{fig:teaser}. 
The detection of visual goals given by exemplars requires to solve a partial matching task, which in essence is a {\it wide-baseline visual correspondence problem}. They are classical in computer vision and at heart of methods in visual localization and relative pose estimation~\citep{visloc22,R2D22019,sarlin20superglue}. We argue that in navigation, however, they did not get the attention they deserve. They tend not to be identified nor solved as such. 

\begin{wrapfigure}{r}{4cm} \centering
\vspace*{-3mm}
\includegraphics[width=4cm]{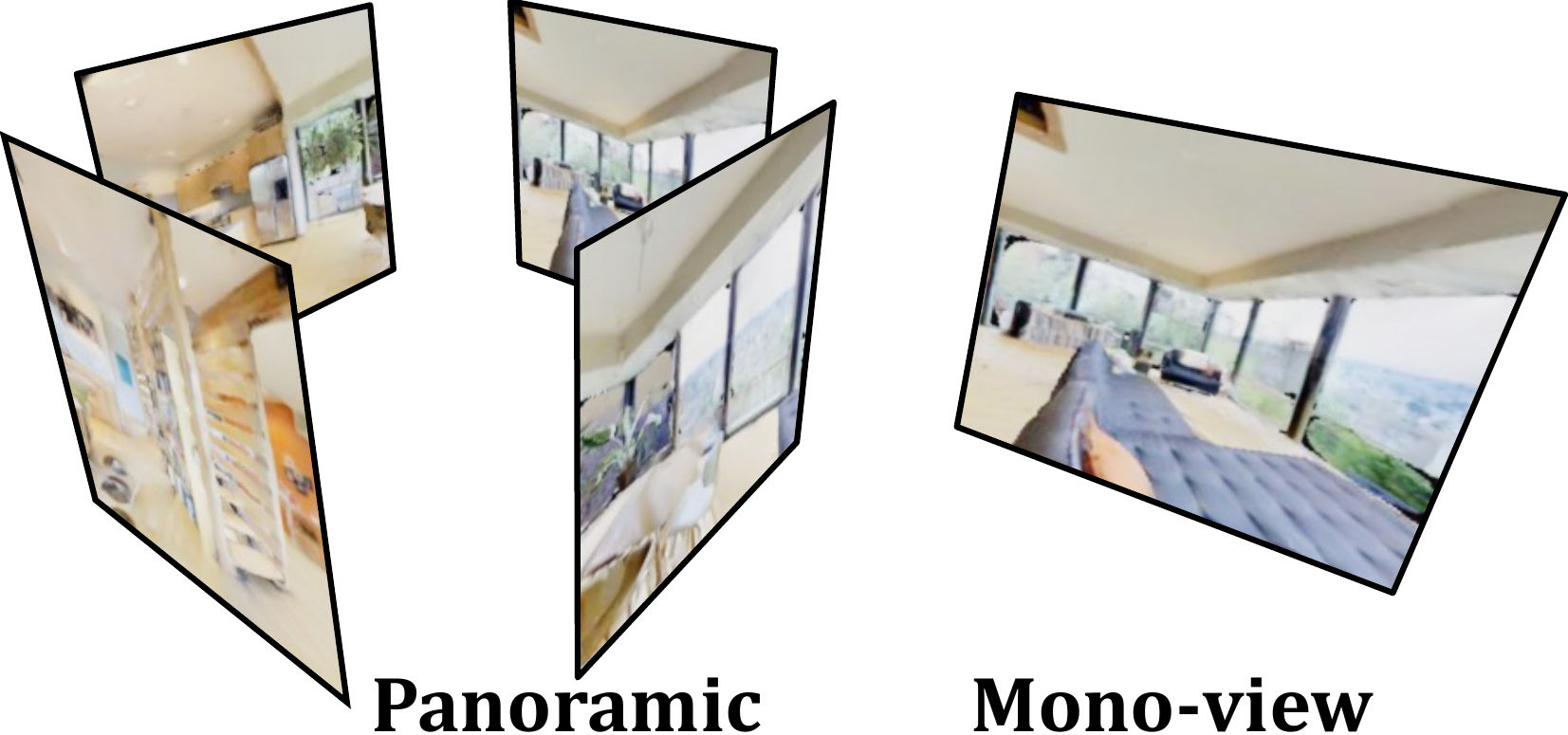}
\caption{\label{fig:pano2mono}Panoramic vs. mono-view input.}
\vspace*{-3mm}
\end{wrapfigure}
Instead, robot perception is addressed through scene reconstruction, for instance with SLAM \citep{Chaplot2020Learning,lluvia_active_2021,thrun2005probabilistic} or by putting the full burden of perception on a visual encoder trained end-to-end from objectives like RL~\citep{DBLP:conf/iclr/JaderbergMCSLSK17, zhu_target-driven_2017} or imitation learning~\citep{DBLP:conf/nips/DingFAP19}. The former does not address goal detection, which needs to be outsourced to an external component. The latter, when trained on tasks like \textit{ImageNav}, attempts to solve the problem implicitly without direct supervision through weak learning signals. This has been shown to be difficult, witnessed by the wide usage of more complex sensors for the \textit{ImageNav} task compared to tasks like \textit{ObjectNav}, where the goal is specified through its category. 
For several years, state-of-the art methods for \textit{ImageNav} used panoramic images consisting of 4 observed images taken at angles of 90° (see Figure \ref{fig:pano2mono}), which facilitates learning the underlying pose estimation task from weak learning signals, but is restrictive in terms of robotic applications. Only recently the field switched to mono-view input.

We propose a new method for end-to-end training of goal oriented visual navigation, which introduces a sequence of pretext tasks of directional learning and visual correspondence and uses them as navigational insights.
We take advantage of recent advances in unsupervised model pre-training for low-level scene understanding and build on the work of~\citet{CroCo2022}, which proposes a type of multi-view pretext task, namely {\it cross-view completion}, a 3D variant of masked image modeling.
We show that the underlying correspondence problem solved by this model is particularly relevant to the \textit{ImageNav} problem and adapt this model through a new dual-encoder model.

In recent work~\citep{krantz2023navigating}, the same problem is addressed with explicit feature matching combined with a modular map-based method. In contrast, our method does not rely on explicit correspondence calculations, the entire agent is differentiable and the perception module is comprised of a combination of a monocular and a binocular encoder. We do, however, show that correspondence solutions emerge from pre-training through the cross-attention behavior, see Figure~\ref{fig:visattention} in the experimental section. 

We present the following contributions:
(i) We define a pretext task of extremely wide-baseline relative pose estimation highly correlated with navigation and positioning cues from visual input. We also introduce a new dataset extracted from the Gibson dataset~\citep{xia2018gibson} tailored for navigation; 
(ii) We couple relative pose estimation with the estimation of visibility, which we link to the specific capacity to decide whether to explore or to exploit;
(iii) We additionally perform self-supervised pre-training for cross-view completion (``\textit{CroCo}'')~\citep{CroCo2022} and show its impact;
(iv) We show that correspondence solutions emerge from pre-training with these tasks.
(v) We propose a dual visual-encoder architecture based on vision transformers and cross-attention, which we integrate into an end-to-end agent and, as proof-of-concept, into a modular architecture;
(vi) We obtain SOTA performance on two standard benchmarks, \textit{ImageNav} and \textit{Instance-ImageNav}.

\section{Related Work}
\label{sec:relatedwork}

\myparagraph{Visual navigation} navigation has been classically solved in robotics using mapping and planning~\citep{burgard1998interactive,macenski2020marathon,marder2010office}, which requires solutions for mapping and localization~\citep{bresson2017simultaneous, labbe19rtabmap,thrun2005probabilistic}, 
for planning~\citep{konolige2000gradient, sethian1996fast} and for low-level control \citep{fox1997dynamic,rosmann2015timed}. These methods depend on accurate sensor models, filtering, dynamical models and optimization. End-to-end trained models directly map input to actions and are typically trained with RL~\citep{DBLP:conf/iclr/JaderbergMCSLSK17,mirowski17learning,zhu_target-driven_2017} or imitation learning~\citep{DBLP:conf/nips/DingFAP19}. They learn
representations, either flat recurrent states or occupancy maps~\citep{Chaplot2020Learning}, semantic maps~\citep{chaplot2020object}, latent metric maps~\citep{DBLP:conf/pkdd/BeechingD0020,Henriques_2018_CVPR,DBLP:conf/iclr/ParisottoS18}; topological maps~\citep{BeechingECCV2020,Chaplot_2020_CVPR,shah_viking_2022}, self-attention~\citep{chen_think_2022,du_vtnet_2021,Fang_2019_CVPR,reed_generalist_2022} or implicit representations~\citep{Marza2022NERF}. Our method is end-to-end trained but adds pretext tasks for perception.

\myparagraph{Goal-oriented navigation}
\label{ssec:perception}
In the easier \textit{ObjectNav} setting, the goal is provided as a category and a detector can encode object shapes in model parameters, trained explicitly for detection, e.g. with semantic  maps~\citep{chaplot2020object}, map-less object detectors~\citep{Savva_2019_ICCV} or image segmenters~\citep{maksymets21thda}, or end-to-end through the navigation loss. \textit{ImageNav} provides the goal as an exemplar image and is a significantly harder task, requiring the perception model to learn a matching strategy itself. Most work are based on end-to-end training~\citep{zhu_target-driven_2017,MezghaniMemAug2021,majumdar2023search}, potentially supported through self-supervised losses~\citep{majumdar2022}. Modular approaches have also been proposed~\citep{DasNeuralModularControl2018,wu22image_goal}. The very recently introduced \textit{Instance-ImageNav} task requires to handle different camera intrinsics and heights between observation and goal, which prior work does with explicit feature matching \citep{krantz2023navigating}. We make it possible to address the \textit{ImageNav} and \textit{Instance-ImageNav} with end-to-end methods in the challenging mono-view setting through new pretext tasks.
 
\myparagraph{Pretext tasks in CV and navigation}
\label{ssec:pretext}
widely used in NLP and CV, pretext tasks aim at learning representations followed by fine-tuning for particular tasks~\citep{bert,yuan2021florence}.
In navigation or robotics, known forms are depth prediction~\citep{das_embodied_2018,DasNeuralModularControl2018,mirowski17learning}, contrastive self-supervised learning (SSL)~\citep{majumdar2022} or privileged information from the simulator like object categories~\citep{EpisodicTransformersICCV2021}, goal directions~\citep{marza2021teaching} 
or visual correspondence in visuomotor policy~\citep{florence20selfsupervised}.
Supervised learning requires in-domain data collection that makes extension beyond the training environment and task difficult. 
Alternatives come from SSL~\citep{wang2022visual,xie2022pretraining}. 
Recent pre-trained visual encoders, like DINO~\citep{dino21} and masked autoencoders (MAE)~\citep{mae22}, have been used in~\citep{yadav2022offline} and~\citep{yadav2023ovrlv2}, respectively. Once pre-trained, the encoder is often frozen before passing into a policy learning module. To be effective across a range of real-world robotic tasks,~\citet{radosavovic22real} diversified the image sources when pre-training. \citet{MezghaniMemAug2021} favor nearby frames to have similar visual representations.

\myparagraph{Pose}
\label{ssec:sota-rpe}
{\it Relative Pose Estimation} (RPE) 
has been intensively studied in computer vision~\citep{kendall,kim22camera,xu22poseestimation}. 
It evolved from local feature based methods (SURF, ORB) and correspondences~\citep{orbslam} to end-to-end training~\citep{li2021pose} or transfer from large-scale classification~\citep{melekhov2017} and finetuning after pre-training on geometric tasks~\citep{CroCo2022}.
Existing solutions revise various components of the regression pipeline~\citep{jin21widebaselines}, discretize the distribution over poses~\citep{chen21widebaseline}, etc. 

Conventional RPE was developed for rather small camera displacements and assumes high visual overlap between images. {\it Wide-baseline RPE} refers to a challenging scenario of large view-point changes and occlusions, on which not all conventional methods work well~\citep{jin21widebaselines}.
Even more challenging, navigation requires what we call {\it Extremely wide-baseline RPE and Visibility}, as an agent may be located in different or cluttered places and therefore may have small or no visual overlap at all. It also requires detecting overlap / visibility, which we perform in this work.

\section{Learning perception for goal oriented visual navigation}
\label{sec:task}
\label{sec:method}
\noindent
We target image-goal navigation in 3D environments (\textit{ImageNav} and \textit{Instance-ImageNav}), where an agent is asked to navigate from a starting location to a visual goal (Figure~\ref{fig:teaser}). The agent receives at each timestep $t$ a single image observation $\mathbf{x}_t{\in} \mathbb{R}^{3{\times}H{\times}W}$ and a goal image $\mathbf{x}^*{\in} \mathbb{R}^{3{\times}H{\times}W}$, both of size $112{\times}112$.
The agent can select one action from the action set $\mathcal{A}$  =\{{\tt \small MOVE FORWARD 0.25m}, {\tt \small TURN LEFT $10^{\circ}$}, {\tt \small TURN RIGHT  $10^{\circ}$}, and {\tt \small STOP}\}. Navigation is considered successful if the {\tt \small STOP} action is selected when the agent is within 1m of the goal position in terms of geodesic distance. For \textit{Instance-ImageNav}, two additional actions are {\tt \small LOOK UP} and {\tt \small LOOK DOWN}.

\begin{figure}[t] \centering
\includegraphics[width=\linewidth]{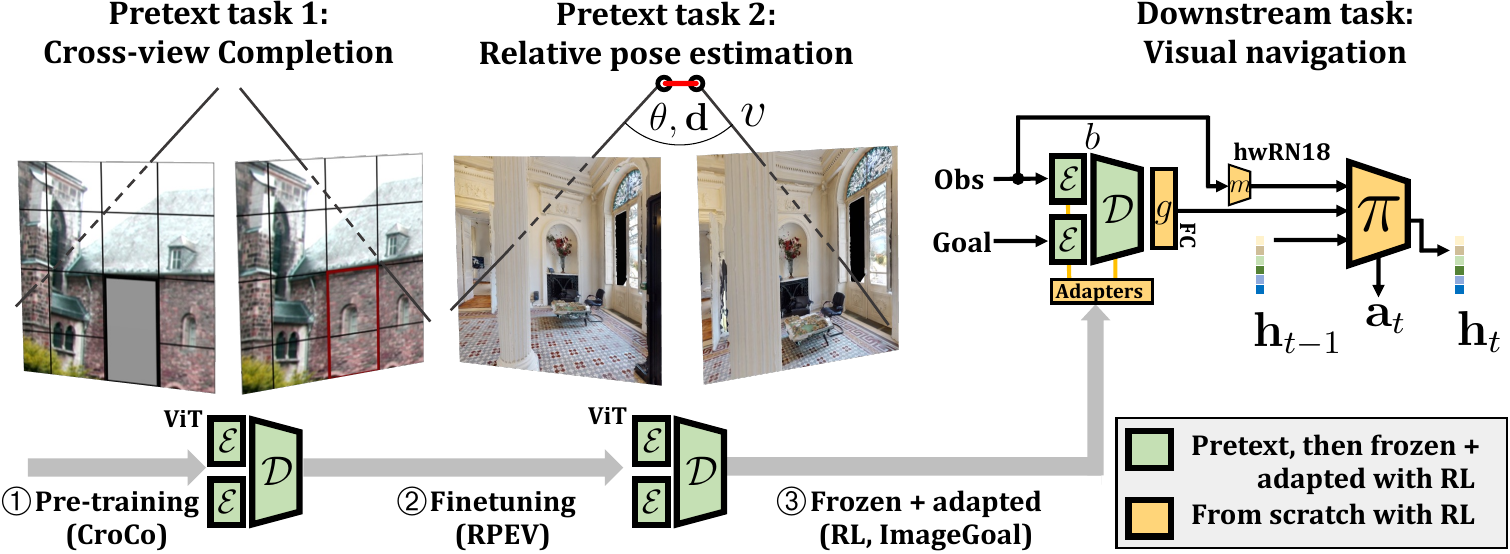} \\[-0.2cm]
\caption{\label{fig:finetuningdistilling}We argue that one of the main bottleneck in goal oriented visual navigation is perception, in particular solving correspondence problems, which we address through two pretext tasks: \ding{192} cross-view completion, introduced by~\citet{CroCo2022}, which reconstructs a masked image from a reference image taken from a different viewpoint, and \ding{193} relative pose and visibility estimation. They are learned by a binocular ViT $b$ and then combined with a monocular encoder $m$ taking only observations, forming the dual encoder DEBiT. The combined predictions are provided to a recurrent policy, maintaining memory $\mathbf{h}_t$ and predicting actions $\mathbf{a}_t$. Monocular encoder $m$ and policy are trained with RL~\ding{194}, the high-capacity model $b$ is frozen but adapted.}
\vspace*{-2mm}
\end{figure}

Our objective is to learn a perception module 
which predicts a latent representation 
given an observation and goal. We conjecture that this requires the following three perception skills: 
\begin{description}[labelindent=4mm,leftmargin=8mm,topsep=1mm,parsep=0mm,labelsep=0.2em]
\item[S1 --- Low-level geometric perception] of the 3D structure of the scene, which includes the detection of navigable space, obstacles, walls and exits, key elements for planning.
\item[S2 --- Perception of semantic categories] is not only required in tasks where object categories are given as goals, which is not the case in the more challenging tasks we target, but is also an additional powerful intermediate cue for other required skills, like geometric perception. Detecting navigable space, for instance, is highly correlated with categories like~\textit{Floor, Wall}.
\item[S3 --- Specific object detection] and relative pose estimation under large viewpoint changes (``extremely wide baseline'') is more difficult than the detection of known object classes and requires to solve a visual correspondence problem, potentially helped by semantic cues.
\end{description}
\noindent
In end-to-end approaches, these skills have been traditionally learned directly from reward or with IL, potentially supported by additional tasks like monocular depth prediction~\citep{das_embodied_2018,DasNeuralModularControl2018,mirowski17learning}, contrastive self-supervised learning~\citep{majumdar2022} or privileged information from the simulator. We argue for a more holistic approach and propose a dual visual encoder combined with a multi-step pre-training strategy. Dubbed ``\textit{DEBiT}'' = \textit{Dual Encoder Binocular Transformer}, it consists of a binocular model $b(\mathbf{x}_t,\mathbf{x}^*)$, which targets skill S3, goal detection and goal pose estimation, and a monocular model $m(\mathbf{x}_t)$, which targets skills S1 and S2 not related to the goal $\mathbf{x}^*$ --- see Figure~\ref{fig:teaser}. The two encoders produce embeddings $\mathbf{e}^b_t$ and $\mathbf{e}^m_t$, respectively, which are integrated into a recurrent policy,
\begin{equation}
    \begin{array}{llp{5mm}l}
    \mathbf{e}^b_t & = g(b(\mathbf{x}_t,\mathbf{x}^*)) && 
    \texttt{\scriptsize // skill S3 - goal direction}\\
    \mathbf{e}^m_t & = m(\mathbf{x}_t) &&
    \texttt{\scriptsize // skills S1, S2}\\
    \mathbf{h}_t & = f(\mathbf{h}_{t-1}, \mathbf{e}^b_t, \mathbf{e}^m_t, l(\mathbf{a}_{t-1})) &&
    \texttt{\scriptsize // recurrent state update (agent mem)}\\
    p(\mathbf{a}_t) & = \pi(\mathbf{h}_t), &&
    \texttt{\scriptsize // policy}\\
    \end{array}
    \label{eq:agent}
\end{equation}
where $g$ is a fully connected layer, $l$ is an embedding function, and $f$ is the update function of a GRU \citep{cho-etal-2014-learning}; for clarity we have omitted the equations of gating functions. The monocular encoder $m(\mathbf{x}_t)$ takes over the navigation skills not related to the goal, we therefore kept it reasonably small and propose a half-width ResNet-18 architecture \citep{he2016deep}, which is trained from scratch and from reward.

The binocular visual encoder $b(\mathbf{x}_t,\mathbf{x}^*)$ decomposes into a Siamese monocular encoder $\mathcal{E}$ applied to each image individually and a binocular decoder $\mathcal{D}$ combining both encoder $\mathcal{E}$ outputs, expressed as 
\begin{equation}
    \mathbf{e}^b_t 
    = g(b(\mathbf{x}_t,\mathbf{x}^*) ) 
    = g(\mathcal{D}(\mathcal{E}(\mathbf{x}_t),\mathcal{E}(\mathbf{x}^*))).  
\end{equation}
DEBiT is implemented as a ViT with self-attention layers in both $\mathcal{E}$ and $\mathcal{D}$ and with cross-attention layers in the decoder $\mathcal{D}$. $\mathcal{D}$ can naturally represent the correspondence problems between image patches through the attention distribution, as we will experimentally show in Section \ref{sec:exp}. For more details we refer to~\citep{CroCo2022} and to Appendix \ref{ssec:app-b}. 

Training the large-capacity binocular encoder entirely from scratch through reward in navigation is difficult. The underlying geometric correspondence problem is complex and can't be handled by the weak learning signal, in particular since the navigation policy needs to jointly learn multiple perception skills, plus some form of internal mapping as well as planning. Training perception separately through losses highly correlated to the perception skills we identified above, in particular S3, proved to be a key design choice --- see Figure \ref{fig:finetuningdistilling}: we pre-train the binocular model $b$ with the \textit{CroCo} pretext task requiring reasoning on low-level geometry (Section~\ref{sec:pretextcroco}) and then finetune it on a novel pretext task dedicated to \textit{ImageNav} (Section~\ref{ssec:rpe}).

\subsection{Cross-view completion} 
\label{sec:pretextcroco}
\noindent

Recently,~\citet{CroCo2022} introduced~\textit{Cross-View Completion} (CroCo), a potent pre-training task trained from a large amount of heterogeneous data and which captures the ability to perceive low-level geometric cues highly relevant to vision downstream tasks. 
It is an extension of masked image modeling~\citep{mae22}
processing pairs of images $(\mathbf{x}, \mathbf{x}')$, which correspond to two different views of the same scene with important overlap. The images are split into sets of non-overlapping patches $\mathbf{p}=\{p_i\}_{i=1\mydots{N}}$, and $\mathbf{p}'=\{p'_i\}_{i=1\mydots{N}}$, respectively. 
The first input image $\mathbf{x}$ is partially masked, and the set of non-masked patches is denoted $\tilde{\mathbf{p}}$. The pretext task requires the reconstruction of the masked content $\mathbf{p} \backslash \tilde{\mathbf{p}}$ from the visible content in the second image, therefore during pre-training we replace the final FC layer $g$ by a patch-wise reconstruction layer denoted $r$,
\setlength{\abovedisplayskip}{5pt}
\setlength{\belowdisplayskip}{5pt}
\begin{equation}
\hat{\mathbf{p}}
= r(b(\tilde{\mathbf{p}},\mathbf{p}'))
= r(\mathcal{D}(\mathcal{E}(\mathbf{p}),\mathcal{E}(\mathbf{p}'))).  
\end{equation}
Training minimizes the MSE loss:
\setlength{\abovedisplayskip}{2pt}
\setlength{\belowdisplayskip}{2pt}
\begin{equation}
\mathcal{L}\left(\mathbf{x},\mathbf{x}'\right)=
   \frac{1}{\left|\mathbf{p} \backslash \tilde{\mathbf{p}}\right|} 
   \sum\mathop{}_{\mkern-5mu \mathbf{p}_i \in \mathbf{p} \backslash \tilde{\mathbf{p}}}
   \left\|\hat{\mathbf{p}}_i-\mathbf{p}_i\right\|^2.
\end{equation}
CroCo is applicable to monocular and binocular downstream problems, competitive performance was shown for monocular depth estimation, optical flow and RPE~\citep{CroCo2022}. 

\myparagraph{Model and Data} 
We use the publicly available\footnote{\scriptsize \url{https://github.com/naver/croco}} code from~\cite{CroCo2022}. We retrained the model ourselves and also explored smaller, more robot-friendly variants. Pre-training data consists of 1.8 million image pairs rendered with the Habitat simulator. 

\vspace*{-2mm}
\subsection{Relative pose and visibility estimation in navigation}
\label{ssec:rpe}
\noindent
Once the binocular encoder $b$ is pre-trained with CroCo, we finetune it on a second pretext task, relative pose estimation and visibility (RPEV) for navigation settings. While for navigation purposes only a 2D vector $\mathbf{t}$ is relevant, which encodes the direction and distance from the agent to the goal, we train the prediction of the full classical {\it relative pose estimation} (RPE) problem, which also includes a $3\times 3$ matrix $\mathbf{R}$ representing the relative rotation of the camera capturing the goal image w.r.t. the current agent orientation. While not useful for navigation, it can potentially add useful learning signals.

\myparagraph{Visibility}
Classically, accurate pose estimation assumes that two images (observation and target) share a sufficiently large part of the visual content, with the overlap providing cues sufficient to estimate the translation and rotation components from one image to the other. This assumption was also satisfied in~\citep{CroCo2022}, but it is, by far, not a valid assumption in navigation. The agent is initially placed far from the goal location and is required to explore the scene, in which case the RPE task cannot be solved through geometry and correspondence, as no scene points are shared between the two images. Recent work has shown that regularities in scene layouts can be exploited to predict distributions over unseen object positions with some  success~\citep{PONI2022}, but this has been reported for object categories and it is unsure whether similar results can be achieved for image exemplars.

For this reason, we added a {\it visibility} measure to our training data, which addresses two issues: 
(i) it ensures feasibility of RPE and excludes image pairs with insufficient correspondence from training the translation and rotation pose components. We do, however, train the pose components even for low amounts of overlap and treat low visibility as an extreme case (``extremely wide baseline''); 
(ii) it provides an additional feature to the agent, as visibility is a strong prior in both positive and negative cases. High visibility indicates closeness to the goal, which can be exploited directly through directional information $\mathbf{t}$ provided by the same model, captured in the embedding $\mathbf{e}^b_t$. Low visibility suggests to explore the scene and rather move away from the current position.

Compared to alternatives like frustrum overlap~\citep{baltnas_relocnet_2018}, we define visibility $v{\in}[0,1]$ as the proportion of patches $\mathbf{p}'_i$ of the goal image $\mathbf{x}'$ which are visible in the observed image $\mathbf{x}$. Note, that this definition is not symmetric, and exchanging the two images alters the visibility value. 

\myparagraph{The RPEV model}
we predict the two RPE components, translation $\mathbf{t}{\in}\mathbb{R}^3$ and rotation matrix $\mathbf{R}{\in}\mathbb{R}^{3\times 3}$, as well as visibility $v$, from an additional head $h$ (which is actually composed of three individual heads) attached to the binocular encoder, taking the embedding $\mathbf{e}^b$ as input,
\begin{equation}
(\mathbf{t},\mathbf{R},v) = h(\mathbf{e}^b) = h(b(\mathbf{x},\mathbf{x}^*)),
\label{eq:rpe}
\end{equation}
where $\mathbf{x}$ is the observed image and $\mathbf{x}^*$ is the goal image. 
To ensure that $\mathbf{R}$ is a valid rotation matrix, we use orthogonal Procrustes normalization from the Roma library~\citep{bregier2021deepregression}.

After CroCo pre-training, we finetune the model with the following loss:
\begin{align}
\mathcal{L}_{RPEV} = \sum\mathop{}_{\mkern-5mu i}
  \Bigl[
    |v_i-v_i^*| + \mathbf{1}_{v^*_i{>}\tau}
    \bigl\{
    |\mathbf{t}_i - \mathbf{t}_i^*| + |\mathbf{R}_i - \mathbf{R}_i^*|
    \bigr\}
  \Bigr] ,
\label{eq:loss_full}
\end{align}
where $i$ indexes image pairs, $\mathbf{t}_i^*, \mathbf{R}_i^*, v_i^*$ denote ground truth values, $\mathbf{1_.}$ is the binary indicator function, $|.|$ denotes the $L_1$ loss and $\tau$ is a threshold which switches off RPE supervision in the case of insufficient visibility.

\myparagraph{Dataset}
We collect a dataset tailored to perception in \textit{ImageNav} by sampling random views from scenes in the Gibson~\citep{xia2018gibson}, MP3D~\citep{chang2018matterport3d} and HM3D~\citep{ramakrishnan2021hm3d} datasets. We respect the standard train/val scenes split of each dataset.
We sample two points uniformly on the navigable area and query the simulator for the shortest path from one to the other. To balance the difficulty of the dataset, we split this path into 5 parts corresponding to increasing thresholds on geodesic distance (``in reach'' $\leq 1$m, ``very close'' $\leq 1.5$m, ``close'' $\leq 2$m, ``approaching'' $\leq 4$m, and ``far'' $>4$m), sample $10$ intermediate positions and orientations along the path in each part, from which images are captured. We compute the fractions of pixels from the goal image visible from any of the ones captured along the path using depth frames which are then discarded.
This process is repeated until $100$ trajectories per scene are sampled, yielding a total of near $68.8$M image pairs with position, orientation and visibility labels, representing $140$GB of data.

\myparagraph{Training for navigation}
We train the parameters of the recurrent policy $(f,\pi)$ and the monocular encoder $m$ jointly from scratch with PPO~\citep{schulman2017proximal} with a reward definition in the lines of the one proposed by~\citet{chattopadhyay2021robustnav} for~\textit{PointGoal},
$
r_t=\mathrm{K} \cdot \mathbf{1}_{\text {success}}-\Delta_t^{\mathrm{Geo}}-\lambda,
\label{eq:rewardn}
$
where $K{=}10$, $\Delta_t^{\mathrm{Geo}}$ is the increase in geodesic distance to the goal, and slack cost $\lambda{=}0.01$ encourages efficiency. The binocular encoder is trained in two different variants:
\begin{itemize}[label=-,labelindent=2mm,leftmargin=8mm,labelsep=0.5em,itemsep=0.5mm,nolistsep]
\item \textbf{Frozen} --- we freeze the parameters of the binocular encoder $b$ after the two step pre-training phases (CroCo + RPEV), and then only finetune the FC layer $g$ in equation (\ref{eq:agent}). Faster to train, we will use this configuration for most ablations and analyses in the experimental section.
\item \textbf{Adapted} --- we freeze $b$ as above, but add adapter layers as in AdaptFormers~\citep{chen2022adaptformer}, which are trained with RL jointly with the policy $(f,\pi)$ and $m$. In the next section we will show that this leads to significant performance improvements.
\end{itemize}

\section{Experimental results}
\label{sec:exp}

\myparagraph{Experimental setup} 
We evaluate on both \textit{ImageNav}, where the goal is a random view taken by the camera of the agent, and the more recent \textit{Instance-ImageNav} \citep{krantz2022instanceintroduction}, where the goal depicts a specific object viewed from a different camera. The major parts of the experiments, ablations and analyses are performed on \textit{ImageNav} in the classical setting, as in~\citep{majumdar2022,MezghaniMemAug2021}.
Unless stated otherwise, we trained the models for 200M steps on an A100 GPU. As in prior work, for \textit{ImageNav} we report performance on the 14 Gibson-val scenes and thus use it as a test set, using the unseen episodes provided by~\citet{MezghaniMemAug2021}. For \textit{Instance-ImageNav} we follow the protocol in \citep{krantz2023navigating}.

\myparagraph{Metrics} RPE is evaluated over the pairs with visibility over $\tau$ in the percentage of correct poses for given thresholds on distance and angle, e.g. 1 meter and 10\textdegree. 
Visibility is evaluated over all pairs by its accuracy at $\pm$0.05, i.e., the percentage of prediction within a 0.05 margin of the ground-truth value. Navigation performance is evaluated by success rate (SR), i.e., fraction of episodes terminated within a distance of ${<}1$m to the goal by the agent calling the \texttt{STOP} action, and SPL~\citep{DBLP:journals/corr/abs-1807-06757}, i.e., SR weighted by the optimality of the path,
$
\textit{SPL}=\frac{1}{N} \sum_{i=1}^N S_i \frac{\ell_i^*}{\max (\ell_i, \ell_i^*)} ,
\label{eq:spl}
$
where $S_i$ be a binary success indicator in episode $i$, $\ell_i$ is the agent path length and $\ell_i^*$ the GT path length.

\begin{table}[t] \centering
\caption{\label{tab:capacities}\textbf{Image-Nav: impact of model capacity} of the binocular encoder on RPEV and nav. perf. (CroCo+RPEV, \underline{200M steps of RL}, frozen, no adapters). $L{=}$layers, $H{=}$heads, $d{=}$embedd.dim. }
\setlength{\tabcolsep}{2pt}
\setlength{\aboverulesep}{0pt}
\setlength{\belowrulesep}{0pt}

{\small
\begin{tabular}{Lcccccccrrrrrr}
\toprule
\rowcolor{TableGray2} 
{\textbf{Variant}} & 
\multicolumn{3}{c}{\textbf{\scriptsize Encoder}} & 
\multicolumn{3}{c}{\textbf{\scriptsize Decoder}} & 
{\scriptsize  \textbf{\#params}} &
{\scriptsize  \textbf{Monoc}} &
\multicolumn{2}{c}{\scriptsize  \textbf{\% correct poses}} &
{\scriptsize  \textbf{Vis-acc}} & 
\multicolumn{2}{c}{\scriptsize  \textbf{Nav. perf.}}
\\
\rowcolor{TableGray2}
& 
{\scriptsize \textbf{L}} & 
{\scriptsize \textbf{H}} & 
{\scriptsize \textbf{d}} & 
{\scriptsize \textbf{L}} & 
{\scriptsize \textbf{H}} & 
{\scriptsize \textbf{d}} & 
{\scriptsize \textbf{(binoc)}} & 
&
{\scriptsize \textbf{1m\&10\textdegree}} & {\scriptsize \textbf{2m\&20\textdegree}} & {\scriptsize(\%)} & 
{\scriptsize \textbf{SR} (\%)} & {\scriptsize \textbf{SPL} (\%)} 
\\
\toprule
DEBiT-L {\scriptsize (``Large'')}, no adapters
& {\scriptsize 12} &{\scriptsize 12} & {\scriptsize 768}
& {\scriptsize 8} &{\scriptsize 16} & {\scriptsize 512}
& 120M & {\scriptsize hwRN18}
& {\bf 97.5} & {\bf 98.9} & {\bf 94.0} & 82.0 & {\bf 59.6} 
\\
DEBiT-B {\scriptsize (``Base'')}, no adapters
& {\scriptsize 12} &{\scriptsize 6} & {\scriptsize 384}
& {\scriptsize 8} &{\scriptsize 16} & {\scriptsize 512}
& 55M & {\scriptsize hwRN18}
& 92.5 & 96.8 & 89.3 & {\bf 83.0} & 55.6  
\\
DEBiT-S {\scriptsize (``Small'')}, no adapters
& {\scriptsize 12} &{\scriptsize 6} & {\scriptsize 384}
& {\scriptsize 2} &{\scriptsize 8} & {\scriptsize 256}
& 24M & {\scriptsize hwRN18}
& 82.7 & 93.5 & 81.6 & 79.6 & 52.1 
\\
DEBiT-T {\scriptsize (``Tiny'')}, no adapters
& {\scriptsize 8} &{\scriptsize 6} & {\scriptsize 384}
& {\scriptsize 2} &{\scriptsize 8} & {\scriptsize 256}
& 17M & {\scriptsize hwRN18}
& 80.3 & 92.4 & 80.6 & 79.3 & 50.0 
\\
\bottomrule
\end{tabular}
}

\vspace*{-4mm}
\end{table}

\begin{table}[t] \centering
\caption{\label{tab:pretrain}\textbf{ImageNav: impact of pre-training strategies}: we ablate CroCo and RPEV pre-training. All results on \underline{100M steps of RL only}, frozen, no adapters. \textbf{Left:} training performance curves (SR) for DEBiT-B, best viewed in color. \textbf{Right:} ablated test results.}
\setlength{\tabcolsep}{1pt}
\setlength{\aboverulesep}{0pt}
\setlength{\belowrulesep}{0pt}
\begin{tabular}{ll}
    \begin{minipage}{5cm}    
        \includegraphics[width=4.9cm]{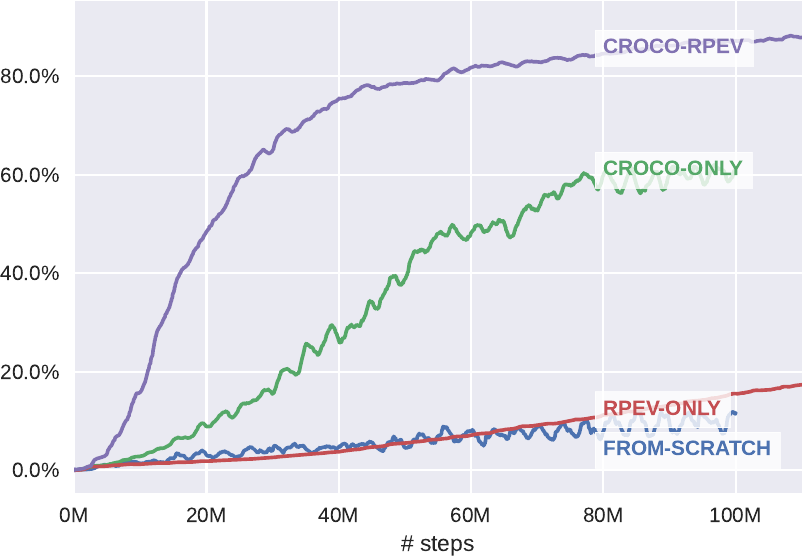}
    \end{minipage}
    &
    \begin{minipage}{8.5cm}
    
{\small
\begin{tabular}{Lccrrrrr}
\toprule
\cellcolor{TableGray2} \textbf{Variant} & 
\multicolumn{2}{c}{\cellcolor{TableGray2} \textbf{Pre-train}} & 
\multicolumn{2}{c}{\cellcolor{TableGray2}\textbf{\% corr. poses}} &
\cellcolor{TableGray2}\textbf{Vis-acc} & 
\multicolumn{2}{c}{\cellcolor{TableGray2}\scriptsize\textbf{Nav. perf.}}
\\
\cellcolor{TableGray2} & 
\cellcolor{TableGray2}{\scriptsize \textbf{CroCo}} & \cellcolor{TableGray2}{\scriptsize \textbf{RPEV}} & 
\cellcolor{TableGray2}{\scriptsize \textbf{1m\&10\textdegree}} & 
\cellcolor{TableGray2}{\scriptsize \textbf{2m\&20\textdegree}} & 
\cellcolor{TableGray2}{\scriptsize(\%)} & 
\cellcolor{TableGray2}{\scriptsize \textbf{SR} } & \cellcolor{TableGray2} {\scriptsize \textbf{SPL} } 
\\
\toprule
DEBiT-L, no adapters & \xmark & \xmark & 
n/a & n/a & n/a & 7.0 & 4.4 \\
DEBiT-L, no adapters & \cmark & \xmark & 
n/a & n/a & n/a & 60.2 & 33.1 \\
DEBiT-L, no adapters & \xmark & \cmark & 
40.1 & 66.7 & 58.3 & 11.8 & 9.9 \\
DEBiT-L, no adapters & \cmark & \cmark & 
{\bf 97.5} & {\bf 98.9} & {\bf 94.0} & {\bf 82.0} & {\bf 54.8} \\
\midrule
DEBiT-B, no adapters & \xmark & \xmark & 
n/a & n/a & n/a & 6.8 & 4.0 \\
DEBiT-B, no adapters & \cmark & \xmark & 
n/a & n/a & n/a & 65.7 & 37.3 \\
DEBiT-B, no adapters & \xmark & \cmark & 
39.7 & 66.4 & 58.8 & 23.6 & 17.4 \\
DEBiT-B, no adapters & \cmark & \cmark & 
{\bf 92.5} & {\bf 96.8} & {\bf 89.3} & {\bf 81.2} & {\bf 53.0} \\
\bottomrule
\end{tabular}
}
    
    \end{minipage}
\end{tabular}
\vspace*{-5mm}
\end{table}

\myparagraph{Baselines} we compare with the state-of-the-art methods on this task, including several variants of \textbf{Siamese Encoders}, which encode the $\mathbf{x}_t$ and $\mathbf{x}^*$ separately, used by a trained policy, typically a recurrent one. First introduced by~\citet{zhu_target-driven_2017}, they were updated by including augmented memory \citep{MezghaniMemAug2021} and powerful ViT based architectures and self-supervised pre-training \citep{majumdar2022,majumdar2023search,yadav2023ovrlv2}. We also compare to the feature-matching based method presented in \citep{krantz2023navigating}, which holds the current SOTA on \textit{Instance-ImageNav}.

\myparagraph{Impact of model capacity} we explore variations in model capacity distributed over the encoder $\mathcal{E}$ and the decoder $\mathcal{D}$ of the binocular visual encoder $b$ (the monocular part $m$ is unchanged) and introduce four different model sizes in Table~\ref{tab:capacities}: DEBiT-L (``\textit{Large}''), DEBiT-B (``\textit{Base}''), DEBiT-S (``\textit{Small}'') and DEBiT-T (``\textit{Tiny}''), where DEBiT-L corresponds to the architecture in~\citep{CroCo2022}. Performance generally improves with more model capacity.

\myparagraph{Impact of pre-training strategies} 
Table~\ref{tab:pretrain} gives results comparing different pre-training strategies for the two largest variants, DEBiT-L and DEBIT-B. Directly training the binocular encoder $b$ from scratch did not lead to exploitable results, reward as a learning signal is too weak. 
CroCo pre-training is essential, directly training on RPEV led to low performance. 
CroCo pre-training alone is not optimal, RPEV adds a significant boost to the gain provided by self-supervised objective alone. The curves in Table~\ref{tab:capacities} (left) shows the evolution of navigation performance (SR) during training, indicating the significant gain and head start the two pretext tasks provide.

\begin{table}[t] \centering
\caption{\label{fig:targetdesign}\textbf{ImageNav: aligning architecture design choices with learning signals}: when both are trained from scratch on navigation reward alone, the Siamese visual encoder~\citep{MezghaniMemAug2021,zhu_target-driven_2017} performs better than our DEBiT architecture. However, DEBiT shines with self-supervised pre-training and fine-tuning, and learning signals which enable learning the correspondence problem solved by the encoder-decoder structure of the binocular stream. RPEV pre-trained models have been added a monocular encoder. Frozen, no adapters.}

\setlength{\tabcolsep}{2pt}
\setlength{\aboverulesep}{0pt}
\setlength{\belowrulesep}{0pt}
\hspace*{-25mm}
\begin{tabular}{cc}
    \begin{minipage}{4.1cm}        
        \includegraphics[width=4cm]{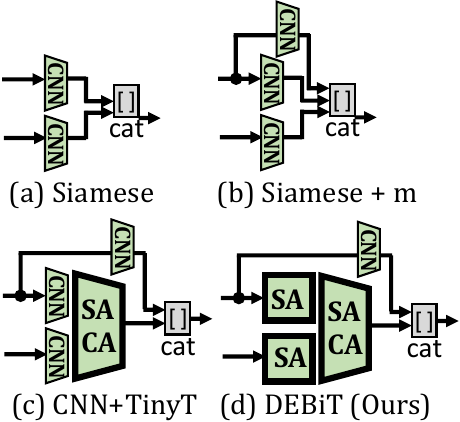}
    \end{minipage}
    &
    \begin{minipage}{7cm}
        \setlength{\tabcolsep}{2pt}        
        
{\small
\begin{tabular}{LLccrr}        
    \toprule    
    \multicolumn{2}{L}{\textbf{Visual encoder}} & 
    \cellcolor{TableGray2}
    \textbf{Pre-train} &
    \cellcolor{TableGray2}
    \textbf{\#parms} &
    \cellcolor{TableGray2}
    \scriptsize \textbf{SR} & 
    \cellcolor{TableGray2}
    \scriptsize \textbf{SPL} \\
    \midrule
    (a) & Siamese hwRN18$^*$     & No                       & 4.1M &      10.1  &       9.6 \\
    (b) & Siamese hwRN18$^*$+$m$ & RPEV                     & 8.3M &       8.0  &       7.7 \\
    (c) & hwRN18+Tiny-T+$m$    & No                       & 10M  &       7.4  &       4.7 \\ 
    (c) & hwRN18+Tiny-T+$m$    & RPEV                     & 10M  &       7.4  &       7.2 \\
    (d) & DEBiT-B (Ours), no adapters   & No                       & 60M  &       6.8  &       4.0 \\ 
    (d) & DEBiT-B (Ours), no adapters   & {\scriptsize CroCo+RPEV} & 60M  & {\bf 83.0} & {\bf 55.6} \\ 
    \bottomrule
    \multicolumn{5}{l}{\scriptsize $^*$ \itshape Baseline in \citep{MezghaniMemAug2021}, inspired by \citep{zhu_target-driven_2017}} \\
    \bottomrule
\end{tabular}
}

    \end{minipage}
\end{tabular}
\end{table}

\begin{table}[t] \centering
{\small
\setlength{\tabcolsep}{2pt}
\setlength{\aboverulesep}{0pt}
\setlength{\belowrulesep}{0pt}
\caption{\label{table:baselines}\textbf{ImageNav: comparisons with prior work}: we gain +12p in SR, +12p in SPL by using RL-trained adapters of the DEBiT encoder. ANS models+weights are from~\citep{Chaplot2020Learning}.}

{\small 
\begin{tabular}{Lcccl}
\toprule
\rowcolor{TableGray2}
\textbf{Method}                & {\bf \#steps} &  {\bf \tiny SR(\%)} & {\bf \tiny SPL(\%)} & {\bf Pretrained weights} 
\\
\midrule
Siam. hwRN18  & 180M &      10.1      &       9.6 & None, from scratch \\ 
Siam. hwRN18 $^2$ & 500M &         -      &       8.0$^1$ & None, from scratch \\
Mem. Aug. {\small \protect\citep{MezghaniMemAug2021}}$^3$    & 500M &         -      &       9.0$^1$ & Finetuned \\
ZSEL      {\small \protect\citep{ZSEL_2022}}        & 500M &      29.2$^1$  &      21.6$^1$ & Obs.\&policy frozen, goal from scratch \\
ZSON      {\small \protect\citep{majumdar22zson}}   & 500M &      36.9$^1$  &      28.0$^1$ & Obs. finetuned, goal frozen (CLIP)\\
VC1-ViT-L {\small \protect\citep{majumdar2023search}}        & 500M &      81.6$^1$  &         - & Finetuned \\
OVRL      {\small \protect\citep{yadav2022offline}}        &  500M &      54.2$^1$  &      27.0$^1$ & Finetuned \\
OVRL-v2   {\small \protect\citep{yadav2023ovrlv2}}        &  500M &      82.0$^1$  &      58.7$^1$ & Finetuned \\
\midrule
ANS \citep{Chaplot2020Learning} + DEBiT-L    &     &      32.0      &      15.0 & Modular architecture + our frozen encoder \\
Ours (DEBiT-B), no adapters                        & 200M &      83.0     &      55.6 & Frozen \\
Ours (DEBiT-L), no adapters                        & 200M &      82.0      &     59.6 & Frozen \\ 
Ours (DEBiT-L) + adapters             & 200M &     \textbf{ 94.0}   & {\bf 71.7} & Frozen + adapted \\
\bottomrule
\multicolumn{5}{l}{\itshape$^1$\itshape Perf. from orig. papers;
~~$^2$ Mono-view ablation of baseline in Table~III of {\small \protect\citep{MezghaniMemAug2021}};} \\
\multicolumn{5}{l}{\itshape$^3$ Retrained in mono-view settings, see Table~1 of {\small \protect\citep{ZSEL_2022}}} 
\\
\bottomrule
 
\end{tabular}
}

}
\vspace*{-4mm}
\end{table}

\myparagraph{Aligning architecture design choices with learning signals} visual encoders for end-to-end trained solutions in the literature for \textit{ImageNav} are typically based on Siamese networks, where the inputs $\mathbf{x}_t$ and $\mathbf{x}^*$ are encoded separately, the respective embeddings are passed to current policies. This late fusion approach allows to train the models from weak reward signals, as the individual encoders learn high-level representations which are compared later in the pipeline. We claim that image comparisons of higher quality can be obtained through early fusion, where images are compared close to input on patch-level. We argue that this leads to a finer visual perception, where correspondence information is encoded in the representation in a more direct way, and provides a more useful signal to the policy. Our experiments shown in Table~\ref{fig:targetdesign} corroborate this claim: 
we compare with a widely used Siamese architecture based on half-width ResNet-18 visual encoders taken from \citep{zhu_target-driven_2017} and reused in \citep{MezghaniMemAug2021}.  DEBiT outperforms them when pre-trained with both pretext tasks, as CroCo pre-training allows correspondence on patch level to emerge (see further below), which leads to accurate pose estimates. Training DEBiT from reward alone is difficult. On the other hand, adding RPEV pre-training to the Siamese architecture is not helpful, the architecture based on late embedding-level fusion cannot exploit this signal.

In an additional experiment we verified whether this difference is explained by the presence of a cross-attention layer easing the computation of correspondences. We designed a hybrid architecture, dubbed (c) in Table~\ref{fig:targetdesign}, which combines convolutional Siamese encoders, implemented as a shared hwResNet18, with a Tiny cross-attention (CA) module with 2 layers, 4 heads and 256 dimensions. Performance is lukewarm, it did not manage to capture the cues provided by the pretext tasks.

\myparagraph{ImageNav, comparison with prior work} Table~\ref{table:baselines} compares the proposed model with prior work. DEBiT largely outperforms the competing methods, including the memory augmented model~\citep{MezghaniMemAug2021}, but also models on large-capacity ViTs like the ``\textit{Visual Cortex}'' model VC1 \citep{majumdar2023search} and OVRL2 \citep{yadav2023ovrlv2}. Both have been pre-trained with masked image encoding, but in a monocular frame-by-frame basis and perform late fusion of observation and goal features, which we argue does not ease learning geometric comparisons. 

\myparagraph{Adapters} adding adapters to DEBIT gains additional 12p of success rate and 12p of SPL, as can be seen in Table~\ref{table:baselines}. For \textit{ImageNav}, it is unlikely that this is explained by improvement of the pose estimation performance through RL finetuning. We conjecture, that the adapters allow to pass richer information through the embedding $\mathbf{e}^b_t$ from the DEBiT to the policy.

\begin{wraptable}{r}{8cm} \centering
\vspace*{-8mm}
{\small
\setlength{\tabcolsep}{1pt}
\setlength{\aboverulesep}{0pt}
\setlength{\belowrulesep}{0pt}
\caption{\label{table:baselinesinstance}\textbf{Instance-ImageNav: adapters} enable specifying goal images with different camera intrinsics and heights compared to the obs. Performance reported on val, max/avg over the last 5 checkpoints.}

{\small 
\begin{tabular}{Lcrcrcl}
\toprule
\rowcolor{TableGray2}
\textbf{Method}                & 
{\bf \#steps} &  
\multicolumn{2}{c}{\bf --- SR (\%)  --- } & 
\multicolumn{2}{c}{\bf --- SPL (\%) --- }
\\
\rowcolor{TableGray2}
&
& {\bf max}
& {\bf avg}
& {\bf max}
& {\bf avg}
\\
\midrule
\citep{krantz2022instanceintroduction}             & 3500M &    5.5 & n/a  & 2.3 & n/a \\
\citep{krantz2023navigating}             & n/a &     56.1 & n/a   & 23.3 & n/a \\
{\footnotesize Ours(DEBiT-L)+adapters}             & 200M &     \textbf{61.1}  & \textbf{59.3}
& {\bf 33.5} & {\bf 32.4}
\\
\bottomrule
\end{tabular}
}

}
\end{wraptable}
\myparagraph{The \textit{Instance-ImageNav} task} In Table~\ref{table:baselinesinstance} we compare with the state-of-the-art in the~\textit{Instance-ImageNav} task, where the goal can be taken with arbitrary camera intrinsics (in particular FOV) and from any camera height, not necessarily the height it is installed on the agent. We trained the agent for a total of 200M steps, 100M of which were done one the \textit{ImageNav} task followed by 100M on \textit{Instance-ImageNav}. As CroCo and RPEV pre-training have been done in \textit{ImageNav} settings (equal intrinsics), adapting DEBiT to this OOD situation was a key design choice, and without adapters performance was actually unexploitable. Non-Siamese adapters (different for obs and goal)  gained around 1p of SR compared to Siamese ones. We outperform the current SOTA method \citep{krantz2023navigating} and show that this task can also be addressed without feature matching.

\myparagraph{Integration into a modular architecture} as a proof of concept, we integrate DEBiT into the well known modular method \textit{Active Neural SLAM}~\citep{Chaplot2020Learning}, which has been designed for exploration and is composed of a high-level policy predicting waypoints, and a low-level policy navigating to the goal. We adapted it to \textit{ImageNav} by adding our binocular encoder $b$ as additional perception module, which switches between (1) navigation towards the predicted goal with the local policy or (2) exploration using the global+local policy, otherwise, see Appendix~\ref{ssec:app-policy} for details. Table~\ref{table:baselines} shows that it is outperformed by our end-to-end variant. It suffers from the fact that pose and visibility estimates are directly given to the policy, whereas the end-to-end trained models benefit from the richer latent embeddings passed from the pen-ultimate layers of the visual encoders.

\begin{figure}[t] \centering
\includegraphics[width=3.4cm]{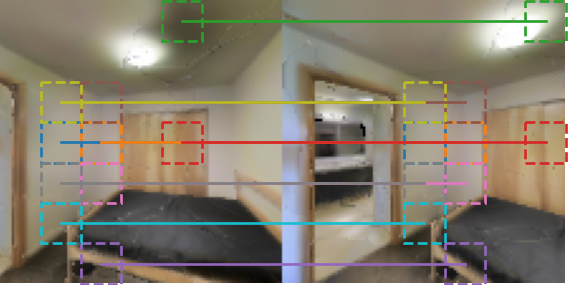}
\includegraphics[width=3.4cm]{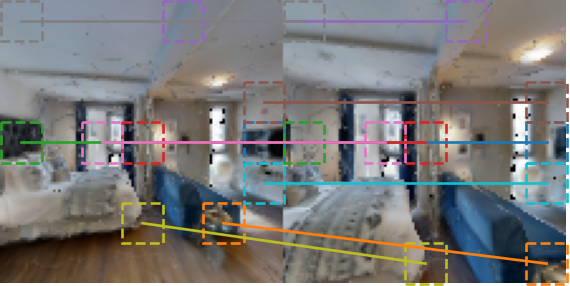}
\includegraphics[width=3.4cm]{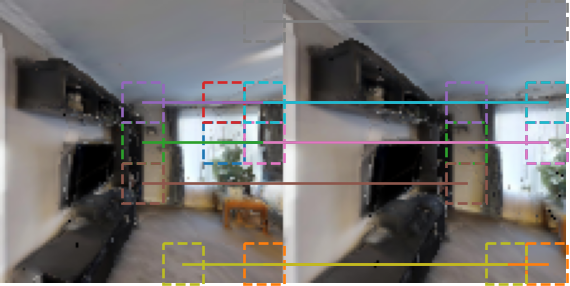}
\includegraphics[width=3.4cm]{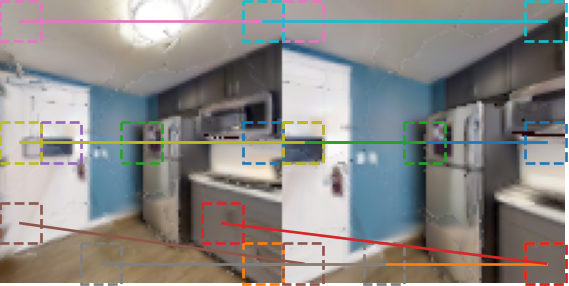}
\includegraphics[width=3.4cm]{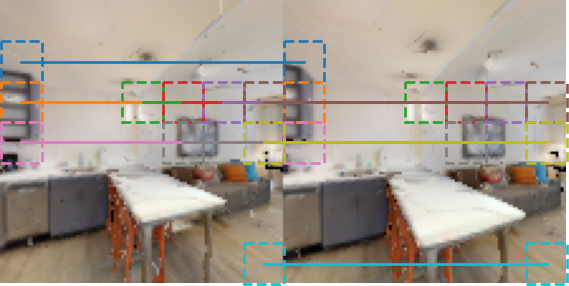}
\includegraphics[width=3.4cm]{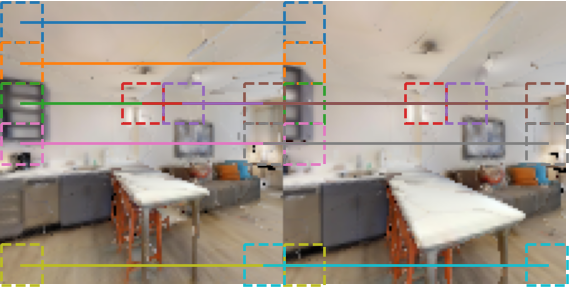}
\includegraphics[width=3.4cm]{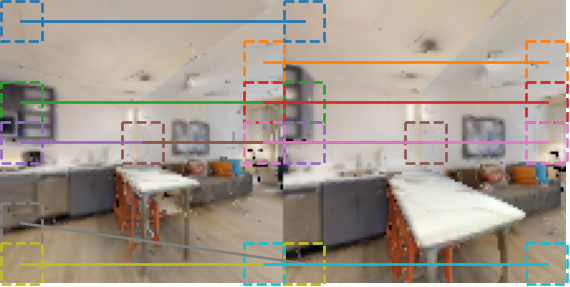}
\includegraphics[width=3.4cm]{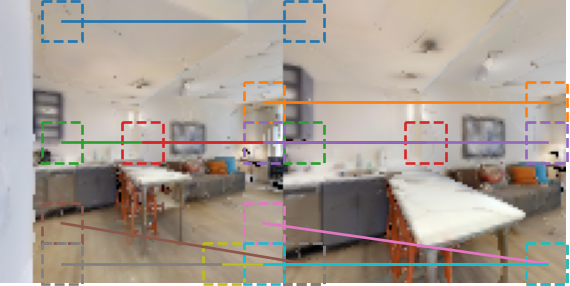}
\\[-0.35cm]
\caption{\label{fig:visattention}\textbf{Emergence of correspondence from pre-training}: visualization of the decoder cross-attention of the finetuned DEBiT-L model for example image pairs. We show attention of the last layer averaged over heads. \textbf{Top:} different scenes and poses. \textbf{Bottom:} image pairs taken from a single trajectory, varying distance to the goal, showcasing robustness to scale changes.}
\vspace*{-3mm}
\end{figure}

\myparagraph{Visualization of attention} in Figure~\ref{fig:visattention} we visualize averaged attention of the last cross-attention layer of a DEBiT-L model. Correspondence solutions naturally emerge without explicit supervision of correspondence solutions. We show a variety of different pairs and poses in the top row, and a single trajectory varying goal distances in the bottom row, indicating robustness to scale changes. 

\begin{figure*}[t] \centering
\includegraphics[width=\linewidth]{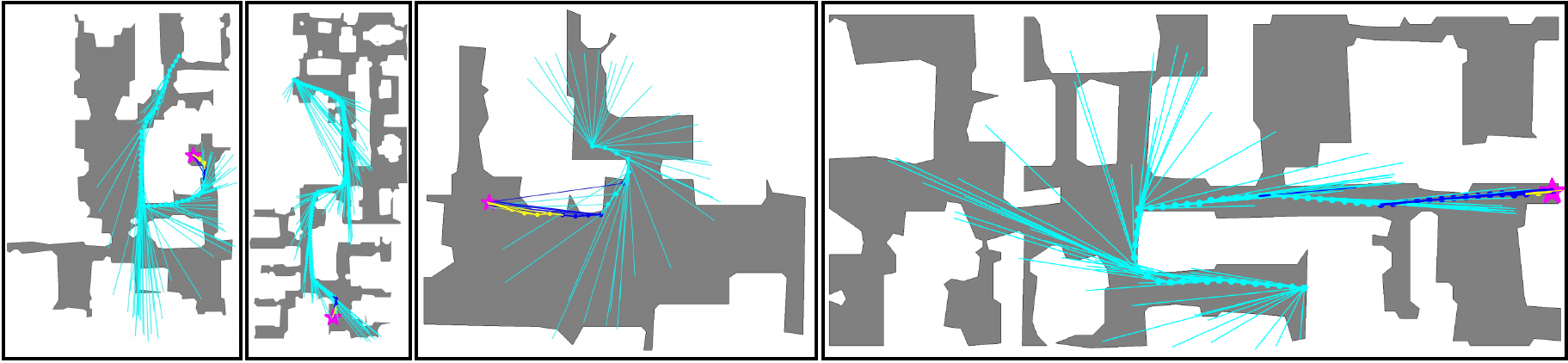} \\[-0.35cm]
\caption{\label{fig:visactions}\textbf{RPEV performance}, DEBiT-L during expert episodes: \textcolor{rpevtarget}{pink star} = goal; segments point to the predicted goal, color encodes visibility prediction $v$ and GT $v^*$: \textcolor{rpevtn}{true negatives (TN) ($v^*{<}\tau, v{<}\tau$)}, \fcolorbox{TableGray2}{TableGray2}{\textcolor{rpevtp}{TP ($v^*{>}\tau, v{>}\tau$)}}, \textcolor{rpevfp}{FP ($v^*{<}\tau, v{>}\tau$)}, \textcolor{orange}{FN ($v^*{>}\tau, v{<}\tau$)} --- rare, not seen. Pose and visibility (and thresholds!) are \textbf{used for visualization only}, the policy receives a latent embedding.}
\vspace*{-4mm}
\end{figure*}

\myparagraph{Visualization of RPEV performance} Figure~\ref{fig:visactions} illustrates pose and visibility estimation performance on several expert trajectories --- DEBiT reliably detects the goal and provides orientations toward it. Let's recall that this information is passed to the policy indirectly through latent embeddings, the RPEV head is discarded after pre-training.

\vspace*{-4mm}
\section{Conclusion}
\vspace*{-2mm}
We have introduced pretext tasks and a dual visual encoder for \textit{ImageNav} and \textit{Instance-ImageNav} navigation in 3D environments, which provide rich geometric information and make it possible to address the challenging mono-view setting with end-to-end trained methods, something the field struggled until now. The method decomposes the problem into several training stages and we show that this makes solutions of correspondence problem emerge without explicit supervision. Experiments show that this leverages the proposed pretext tasks, as well as a dedicated dual encoder architecture. We outperform competing methods and obtain SOTA performance on both benchmarks. We also showcase the integration into a modular navigation pipeline.

\bibliographystyle{iclr2024_conference}
\bibliography{ms}

\appendix

\newpage
\section{Appendix}

All networks have been implemented in PyTorch, below we provide details of the binocular encoder $b$, the monocular encoder $m$, and the recurrent policy composed of a dynamics mapping $f$ and the prediction head $\pi$.

\subsection{The binocular encoder $b$}
\label{ssec:app-b}
DEBiT's binocular encoder $b$ follows the architecture in~\citep{CroCo2022}, and the ``\textit{Large}'' version DEBiT-L is equivalent to \citep{CroCo2022}, corresponding to the code available at {\small \url{https://github.com/naver/croco}}: the encoder $\mathcal{E}$ is a ViT-Base model, i.e., composed of $L=12$ self-attention blocks with $H=12$ heads each and the embedding dimension $d=768$. The decoder $\mathcal{D}$ is composed of $L=8$ cross-attention blocks with $H=16$ heads each and an embedding dimension $d=512$.
We will not detail the encoder part further, as it is very similar to a standard (monocular) ViT. Concerning the cross-attention blocks used in the decoder part, they are composed of the following layers:
\begin{itemize}
    \item A self-attention layer is applied to the features corresponding to the current view (potentially already enriched by features from the goal view in previous blocks), with pre-norm (\texttt{LayerNorm}) and skip-connection.
    \item The resulting features are used as queries on the goal view features (used as keys and values, optionally also pre-normalized) in a cross-attention layer, with skip-connection.
    \item A $2$-layers perceptron with dimensions $d=2048$ and back to $d=512$ independently projects the features of each patch, with pre-norm, GELU activation and skip-connection.
\end{itemize}
Full details on these cross-attention blocks are available in~\citep{CroCo2022}.
Smaller DEBiT versions differ in the number of layers, heads and the embedding sizes, see Table~~\ref{tab:capacities}.

\subsection{The projection $g$}

The decoder blocks are followed by a single patch-wise linear layer $d$, which can also be seen as a $1 \times 1$ convolution on the features of all patches in 2D. It projects them from dimension $512$ to $64$, before flattening them to a $3136$-dim vector (given images of size $112 \times 112$ and patches of size $16\times 16$).

\subsection{Auxiliary heads $h$ for RPEV}

For the second phase of pre-training, the result of the flattened projection $g$ of size $3136$ is ReLU-activated and fed to a common $1024$-dim linear layer, before being dispatched to $3$ independent output layers for the predictions of relative camera translation $\mathbf{t}$ and rotation $\mathbf{R}$, as well as goal visibility from current view $v$:
\begin{itemize}
    \item The translation head directly outputs a 3D vector in the coordinate frame of the current view.
    \item The rotation head outputs a 9D vector which is reshaped as a $3 \times 3$ matrix, constrained to be a valid rotation matrix (using orthogonal Procrustes normalization) with a small regularization term added to the loss.
    \item The goal visibility head linearly outputs a single value with no activation constraining it to be between $0$ and $1$.
\end{itemize}

\subsection{The monocular encoder $m$}

The monocular encoder is a half-width ResNet-18 as frequently used in prior work on visual navigation~\citep{MezghaniMemAug2021,majumdar2022}. 
It is very similar to a standard ResNet-18~\citep{he2016deep}, only differing in $3$ ways:
\begin{enumerate}
    \item Instead of using $64$, $128$, $256$ and $512$ 
    channels in the $4$ layers (of $2$ basic blocks each),
    the half-width ResNet-18 uses $32$, $64$, $128$, and $256$ channels.
    \item All \texttt{BatchNorm2D} layers are replaced by \texttt{GroupNorm} layers with $16$ groups each.
    \item The final (global pooling $+$ linear layer) is replaced by a small ``Compression'' module which consist in:
    a 3x3 convolution (with padding) reducing the number of channels from $256$ to $128$, followed by a \texttt{LayerNorm} and a ReLU activation, whose result is flattened and fed to a linear layer to produce a $512$-dim flat embedding of the current (monocular) view.
\end{enumerate}

\subsection{The recurrent policy $(f,\pi)$}
\label{ssec:app-policy}
The policy relies on a single-layer GRU as our recurrent state encoder.
The $3$ flat features vectors produced by the binocular, monocular, and previous action encoders are concatenated and fed to the GRU, whose output $h_t$ is passed to $2$ linear heads that respectively generates a softmax distribution over the action space (Actor head), and an evaluation of current state (Critic head). This agent is structurally similar to the agent from~\citep{PONI2022,wijmans2019dd}, which is used for the related point-goal navigation task, and achieves 100\% PointGoal success on the Gibson dataset~\citep{xia2018gibson}, but with different input modalities (RGB + ImageGoal instead of RGB-D + PointGoal).

\subsection{The ANS adaptation to \textit{ImageGoal}}

\begin{figure}[ht] \centering
\includegraphics[width=0.9\linewidth]{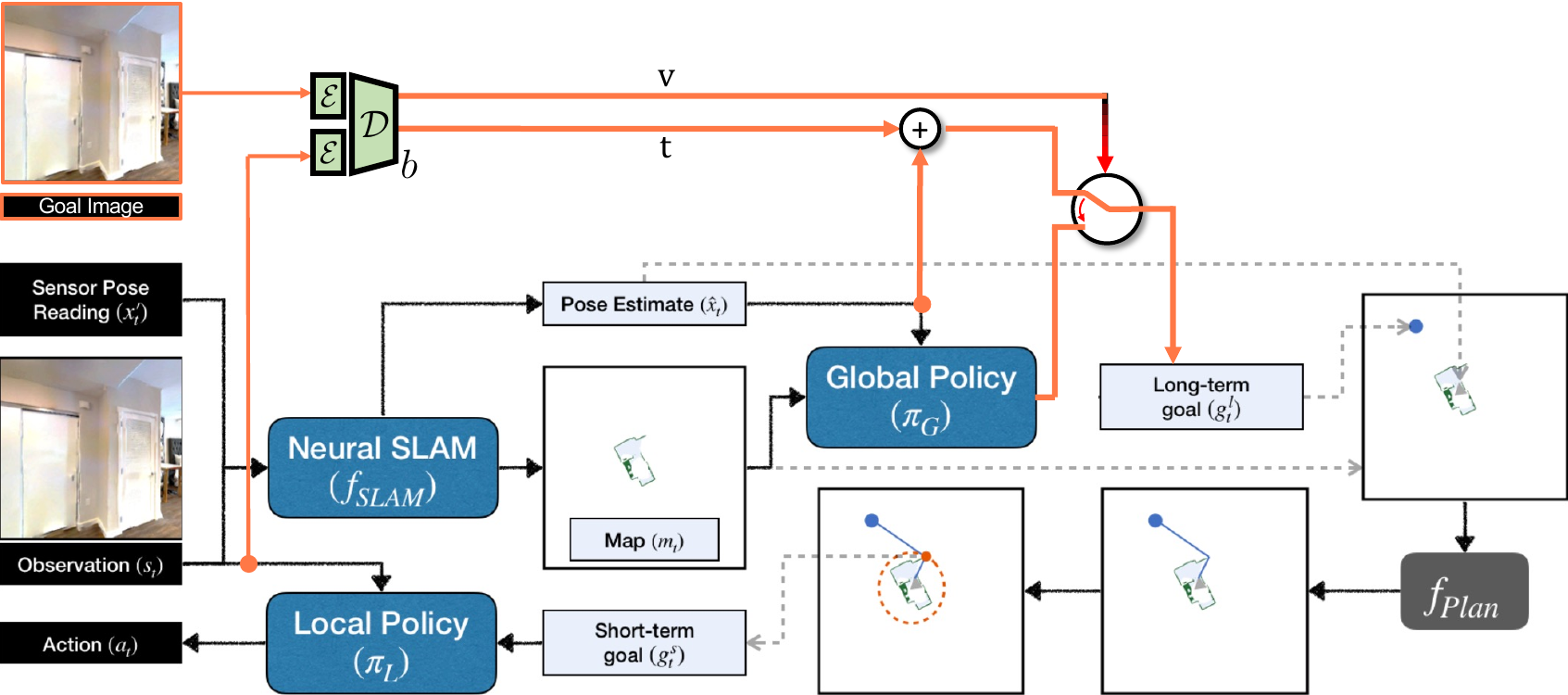}
\caption{Active Neural SLAM + DEBiT-L/$b$ --- we use the binocular encoder $b$ only of the DEBiT architecture. Figure is reproduced from~\citep{Chaplot2020Learning}, with additional parts from our adaptation to \textit{ImageGoal} drawn in~\textcolor{orange}{orange}.}
\label{fig:arch}
\end{figure}

Figure~\ref{fig:arch} shows our adaptation of \textit{Active Neural SLAM}~\citep{Chaplot2020Learning} 
to the \textit{ImageGoal} task. The new components are shown in \textcolor{orange}{orange}. They include the target image, the binocular encoder $b$ as additional perception module, and a module switching between (1) navigation towards the predicted goal with the local policy and (2) exploration using the global+local policy, otherwise. Switching is done by thresholding the visibility prediction $v_t$ with a threshold $T$, whose influence we tested in sensibility study below. 

\end{document}